\documentclass[conference, 10pt, a4paper]{IEEEtran}
\IEEEoverridecommandlockouts
\usepackage{cite}
\usepackage{amsmath,amssymb,amsfonts}
\usepackage{algorithmic}
\usepackage{graphicx}
\usepackage{subcaption}
\usepackage{textcomp}
\usepackage{xcolor}
\usepackage{float}
\usepackage{hyperref}
\hypersetup{
    colorlinks,
    linkcolor={red!50!black},
    citecolor={blue!50!black},
    urlcolor={blue!80!black}
}

\IEEEsettextheight{1in}{1in}

\def\BibTeX{{\rm B\kern-.05em{\sc i\kern-.025em b}\kern-.08em
    T\kern-.1667em\lower.7ex\hbox{E}\kern-.125emX}}
\begin{document}

\newcommand\blfootnote[1]{%
  \begingroup
  \renewcommand\thefootnote{}\footnote{#1}%
  \addtocounter{footnote}{-1}%
  \endgroup
}

\title{Learning Keypoints from Synthetic Data for Robotic Cloth Folding
\thanks{This research is supported by the Research Foundation Flanders
(FWO) under Grant numbers 1S56022N (TL) and 1SD4421N (VDG)}
}

\author{Thomas Lips, Victor-Louis De Gusseme and Francis wyffels \\
\IEEEauthorblockA{\textit{AI and Robotics Lab (AIRO), IDLab,}
\textit{Ghent University - imec}\\
Thomas.Lips@UGent.be}

}
\twocolumn[{%
\renewcommand\twocolumn[1][]{#1}%
\maketitle
 \centering
    \captionsetup{type=figure}

     \begin{subfigure}[b]{0.34\textwidth}
         \centering
         \includegraphics[width=\textwidth]{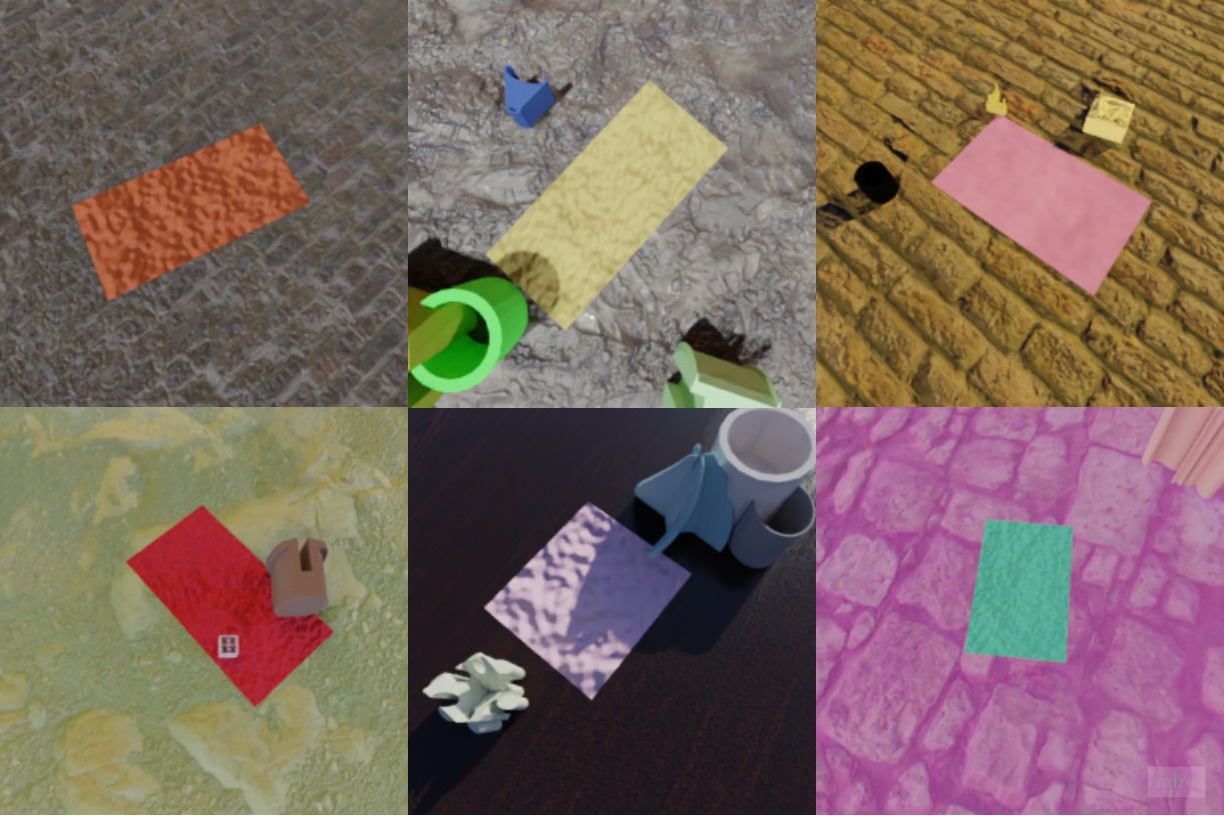}
     \end{subfigure}
     \hfill
     \begin{subfigure}[b]{0.65\textwidth}
         \centering
         \includegraphics[width=\textwidth]{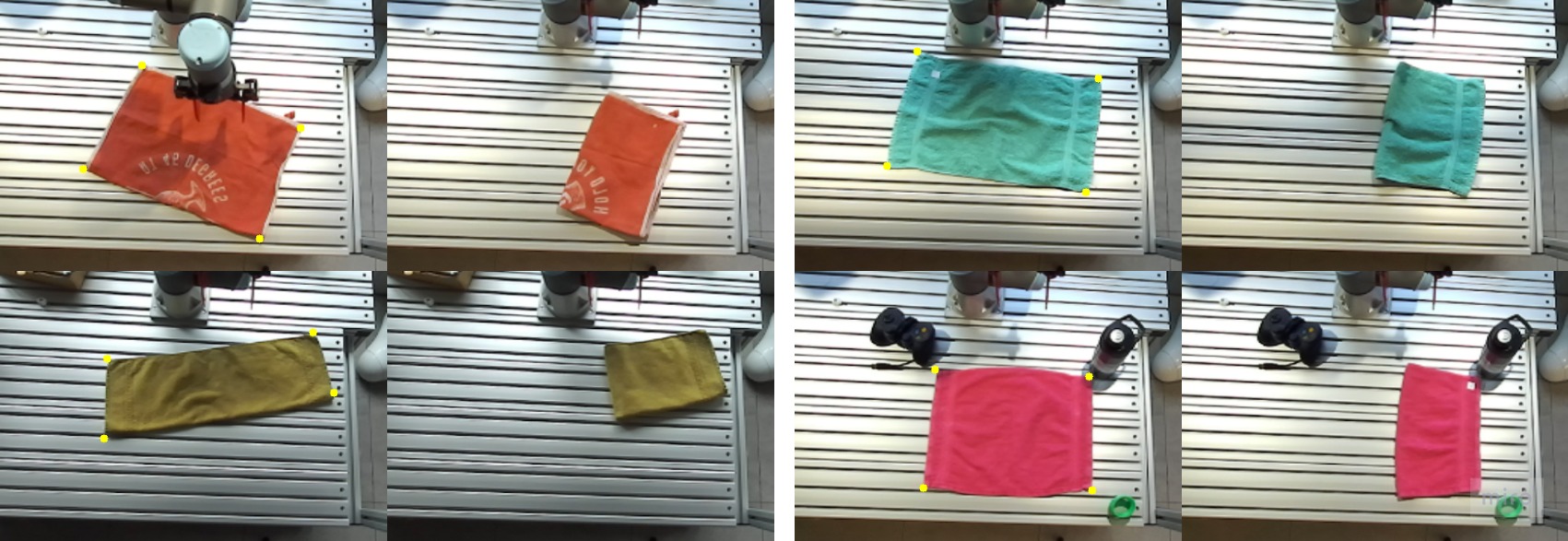}
     \end{subfigure}

        \captionof{figure}{Left - Examples of the synthetic data used to train the keypoint detector. Center/Right - Successful folds on towels from the evaluation set. Each image pair shows the initial state with the detected keypoints (yellow dots) and the final state after attempting to grasp and fold. }
        \label{fig:head-figure}
             \vspace{4mm}

}]

\begin{abstract}
    Robotic cloth manipulation
is challenging due to its deformability, which makes determining its full state infeasible. However, for cloth folding, it suffices to know the position of a few semantic keypoints.
    Convolutional neural networks (CNN) can be used to detect these keypoints, but require large amounts of annotated data, which is expensive to collect.
    To overcome this, we propose to learn these keypoint detectors purely from synthetic data, enabling low-cost data collection. 
    In this paper, we procedurally generate images of towels and use them to train a CNN. We evaluate the performance of this detector for folding towels on a unimanual robot setup and find that the grasp and fold success rates are 77\% and 53\%, respectively.
    We conclude that learning keypoint detectors from synthetic data for cloth folding and related tasks is a promising research direction, discuss some failures and relate them to future work. A video of the system, as well as the codebase, more details on the CNN architecture and the training setup can be found at \url{https://github.com/tlpss/workshop-icra-2022-cloth-keypoints.git}.
\end{abstract}
 

\section{Introduction}
Robotic manipulation of deformable objects is challenging, both in terms of perception, control and modelling \cite{Yin2021}. This is a.o. caused by their high-dimensional state~\cite{sanchez2018deformable_survey}. Nonetheless, any general-purpose robotic manipulation system would encounter such objects. Cloth, for example, is omnipresent in household settings. Although significant progress has been made, cloth manipulation remains a challenging task in terms of perception and control~\cite{Borras2020,Doumanoglou2016,Yin2021,sanchez2018deformable_survey}.

Supervised deep learning (DL) has the potential to tackle these perception tasks and to generalise to a wide range of settings. However, DL is far from data-efficient and requires large numbers of labelled data to learn and generalise. Procedural data generation is an appealing alternative to manually collecting these datasets: Synthetic data is cheap to generate, has perfect annotations and can be used to generate all desired variations. The drawback is that the resulting network often has a reduced performance when transferred to the real world. This is due to the remaining differences between the simulation and the real world, referred to as the reality gap.

In this paper, we tackle towel folding, a standard task for cloth manipulation~\cite{benchmarking_cloth_manipulation, MaitinShepard2010}. 
We approach this task by using a convolutional neural network (CNN) as keypoint detector to estimate the 2D positions of the towel corners from a single RGB image and executing a scripted, open-loop grasp and quasi-static fold trajectory based on these semantic keypoints. We train the keypoint detector entirely on synthetic data and evaluate the zero-shot sim-to-real performance on a unimanual robotic setup under various conditions. Using the terminology of~\cite{verleysen2020folding_demonstrations, Borras2020}, we assume the towels are unfolded but not necessarily perfectly flattened. Recent progress in cloth unfolding has made this a realistic assumption~\cite{ha2022flingbot}. 

Our contributions are as follows:
\begin{itemize}
    \item We show that synthetic data can be used to train convolutional neural networks to detect keypoints for cloth folding.
    \item We extensively evaluate the performance of our keypoint detector and unimanual folding system.
    \item We provide qualitative insights into the failure cases.
\end{itemize}



\section{Related Work}
\label{section:related-work}
\subsection{Robotic Cloth Folding}
Cloth manipulation has been extensively studied~\cite{sanchez2018deformable_survey, Borras2020}. For cloth folding in particular,~\cite{Miller2012, Doumanoglou2016, MaitinShepard2010} have devised complete pipelines.
These works rely  on segmentation and a combination of polygonal approximations and template matching or corner detectors to localise keypoints (landmarks) of unfolded cloth items and subsequently fold them. Although these systems perform very well in their test settings, the segmentation algorithms assume a known background color distribution, which strongly limits the potential to generalise to diverse environments. Furthermore the perception pipeline can take up to 5 seconds~\cite{Doumanoglou2016} to detect the keypoints.



\subsection{Learning Keypoint Representations for Robotic Control}
Recently, CNNs were trained to detect~\cite{vecerik2021s3k, Qin_keypoints_tool_manipulation, wang_door_handle_keypoints, kendall2015posenet} and even discover~\cite{kulkarni2019unsupervised,chen2021unsupervised } semantic keypoints end-to-end for various robotic tasks. As the number of detected keypoints can often vary, keypoint detectors usually output spatial heatmaps instead of cartesian coordinates \cite{jakab_unsupervised_landmarks,zhou2019centernet}.

\subsection{Synthetic Data for Computer Vision}
Procedural data generation for training perception models has been used to learn state representations for various robotic tasks~\cite{Qin_keypoints_tool_manipulation, wang_door_handle_keypoints, tremblay2018sim2real_pose_estimation}. The main limitation is the induced reality gap, which limits the performance on the real-world target domain. Tobin et al.~\cite{tobin2017domainrandomization} introduced domain randomization as a way to overcome the reality gap by enlarging the distribution from which the data is generated to ensure it entails the real-world data. However, more recent work has stressed the importance of more realistic image synthesis~\cite{tsirikoglou2017procedural_physically_automotive, tremblay2018sim2real_pose_estimation} and showed that over-randomizing can result in loss of context and performance~\cite{prakash2019structuredDR}.  
Most authors find that even with domain randomization, the mismatch between both distributions is still too large and resort to finetuning with real-world data to increase performance~\cite{tremblay2018sim2real}. The factors that cause this reality gap are not well-known and one usually attempts to close it with task-specific tuning~\cite{tremblay2018sim2real, tremblay2018sim2real_pose_estimation}.


\section{Method}
\label{section:method}
\subsection{Synthetic Dataset Generation}
\label{subsection:procedural data generation}
We use Blender~\cite{blender} and BlenderProc~\cite{denninger2019blenderproc} to generate data samples. For each image, we build a new 3D scene and randomize object geometries, materials, lighting, and camera pose. Examples can be found in Fig.~\ref{fig:head-figure}.
The towel geometry is modelled as a rectangular mesh. We created a procedural material that combines a random HSV colour and a Perlin noise texture to omit the need for realistic fabric textures, which are difficult to obtain. Additionally, we sample up to 5 distractor objects from a subset of the Thingi10k dataset~\cite{zhou2016thingi10k} and add them randomly to the scene, as this was observed to reduce false positives on the manipulator and other objects present in the scene.
The ground plane material combines a random HSV base colour with a texture from PolyHaven\footnote{\url{https://polyhaven.com/}} to introduce spatial patterns.
To generate complex and realistic lighting of the scene, we use 360-degree images as environment textures, which are also obtained from PolyHaven. The position of the camera is sampled inside a spherical cap and its orientation is set to point towards the centre of the scene. Finally, the scene is rendered into a 256x256 image with Cycles, Blender's physically-based path tracer.
 
\subsection{Keypoint Detector}
To detect the desired keypoints on an RGB image, we use a fully convolutional neural network to predict a single spatial heatmap that contains all corners of the cloth. Inspired by Vecerik et Al.~\cite{vecerik2021s3k}, a U-net architecture~\cite{ronneberger2015unet} is used to combine spatial resolution conserving paths with downsampling paths to obtain a large receptive field.
In the spatial bottleneck, ResNet-inspired skip connections \cite{he2016resnet} are used. Bilinear upsampling is used for the upsampling layers. All hidden layers use the ReLU activation function. The final layer uses a sigmoid, of which the outputs are interpreted as the probability function of having a keypoint centred on that location~\cite{jakab_unsupervised_landmarks}.
Heatmaps for the synthetic data samples are generated using a pixel-wise maximum over 2D Gaussian blobs that are centred on each keypoint~\cite{zhou2019centernet}. Pixel-wise Binary Cross Entropy is used as a training objective and keypoints are extracted using a max-filter with a configurable receptive field, as implemented in Scikit-Image~\cite{van2014scikit}.  We use Pytorch Lightning~\cite{Falcon_PyTorch_Lightning_2019} to train the CNN and Weights and Biases~\cite{wandb} for experiment tracking.


\subsection{Robotic Folding}
To initiate a fold, a top-down RGB image is passed through the keypoint detector network. If less than four keypoints are detected, the folding is aborted; otherwise, the four keypoints with the highest probability are extracted and reprojected onto the table plane.  Based on these keypoints, a local frame is defined and the scripted pregrasp pose, grasp pose and fold trajectory in this local frame are transformed to the robot frame. The sequence is then executed by the robot. The robot grasps the towel in the middle  of the side that needs to be folded. The folding trajectory is an arc from that grasp point to the corresponding point on the other side of the towel. 



\section{Experiments}
\label{section:experiments}
\subsection{Evaluation}
To evaluate the keypoint detector, two sets of towels are used. 
All towels were collected by asking a number of lab members to bring random towels to reduce bias. The first set contains 11 towels with various colours and material properties but with uniform textures, as modelled in the synthetic data. These are referred to as the in-distribution towels. The second set consists of 9 towels that are not in the distribution of the synthetic data as they have highly non-uniform textures or very different material properties. We refer to this additional set as the out-of-distribution towels and include them to evaluate to what degree the neural network is able to generalise from the synthetic data.

We use three different environment settings, in which we vary lighting and the presence of 6 random chosen distractor objects. For each setting, the robot attempts to fold each towel twice. The pose of the towel is manually randomised within the workspace of the robot before each attempt. For some larger towels, we already partially fold the towel to reduce their dimensions. The partially-folded towels range from 0.2\;m to 0.5\;m in size.

As the most informative metric of a robotic system is the task performance, we evaluate the keypoint detector by measuring the grasp success rate. Grasp success is defined as the ability of the system to enclose the cloth at the grasp pose. Note that this is different from~\cite{benchmarking_cloth_manipulation}, where grasps are only considered successful if they are held during the entire manipulation. Fold success is defined, as suggested in~\cite{benchmarking_cloth_manipulation}, as the approximate coincidence of the opposing corners after manipulation. 

\subsection{Keypoint Detector}
Using the procedure described in Section~\ref{subsection:procedural data generation}, a dataset of 30,000 images is generated, of which samples can be found in Fig.~\ref{fig:head-figure}. The dataset is generated on a Dell XPS 9570 laptop with a low-range Nvidia GTX 1050Ti mobile GPU. Generating a single sample takes 1.8\;s on average, of which 0.6\;s is spent on building the scene and 1.2\;s on rendering.

We then train our keypoint detector to predict all visible corner keypoints using sensible hyperparameters based on previous experience. We train for 15 epochs, which takes 55 minutes on an Nvidia V100 GPU and results in an average precision (with 2 pixel threshold) of 80\% on a synthetic validation set. We refer to the accompanying Github page for details about the network architecture and other training parameters.

\subsection{Robotic Setup}
We use a UR3e robot and Robotiq 2F-85 gripper. We 3D-printed fingertips with a width of 0.08\;m using flexfill TPU 98A filament to ensure compliance of the fingers while sliding underneath the cloth for grasping. The camera is a ZED2i that is mounted 1\;m above the table and of which the extrinsics have been determined upfront. The images are cropped to the relevant area and then rescaled to the CNN's input size of 256x256 pixels.


\begin{figure}
    \vspace*{-1.5mm}

     \centering
     \begin{subfigure}[b]{0.49\linewidth}
         \centering
         \includegraphics[width=\textwidth]{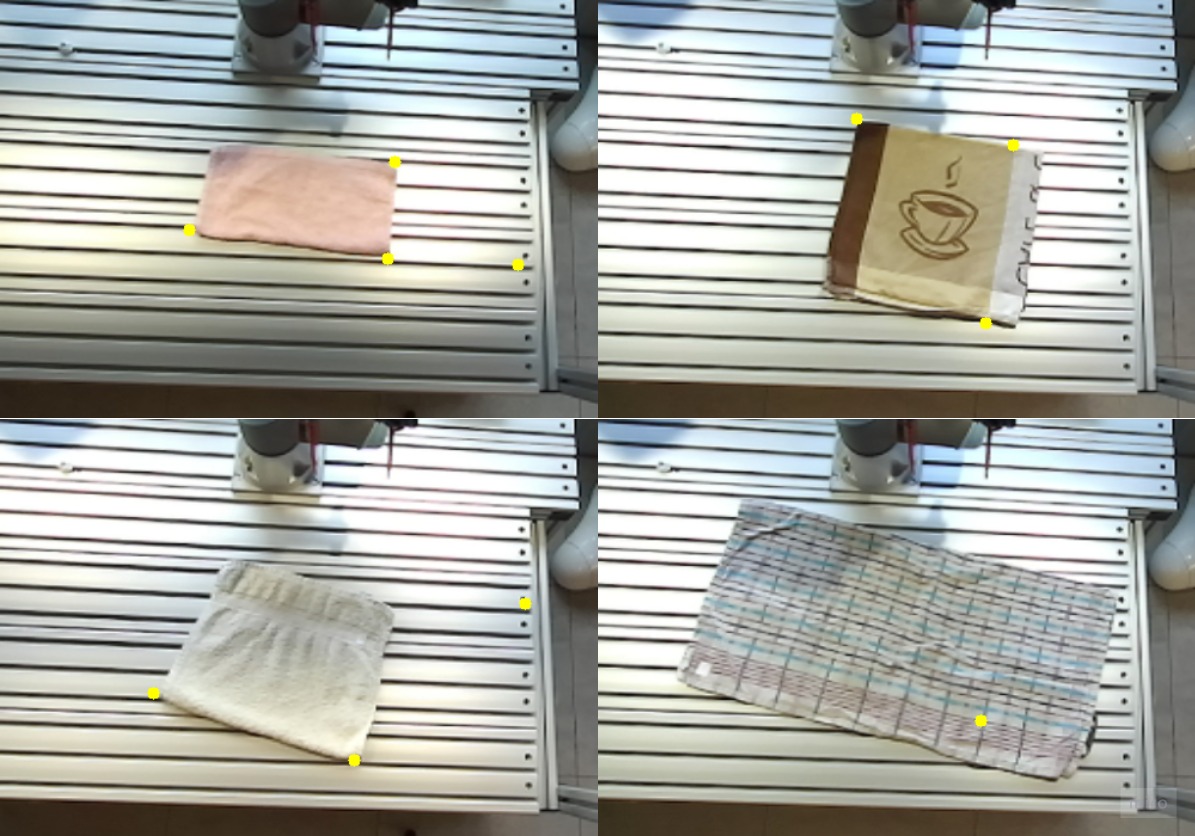}
         \caption{}
         \label{fig:failed-keypoints}
     \end{subfigure}
     \hfill
     \begin{subfigure}[b]{0.49\linewidth}
         \centering
         \includegraphics[width=\textwidth]{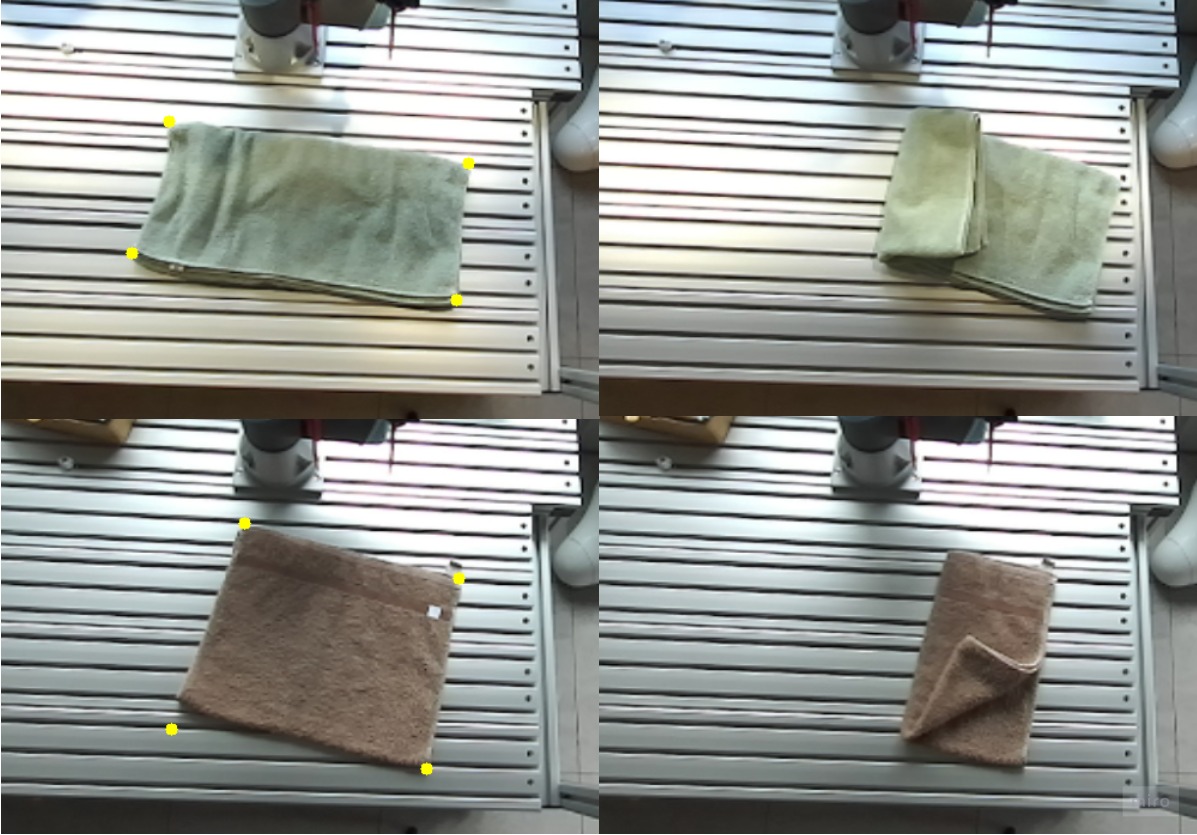}
         \caption{}
         \label{fig:failed-fold-examples}
     \end{subfigure}
        \caption{a) Examples of trials for which the keypoint detection failed, resulting in a failed grasp and fold. b) Examples of trials for which the fold failed after a successful grasp.  Each image pair shows the initial state with the detected keypoints (yellow dots) and the final state after attempting to grasp and fold. }
\end{figure}

\subsection{Results}

The performance of the system is reported in Table~\ref{table:evaluation-metrics}. The total grasp success rate for the towels that are similar to those modelled in the synthetic data is about 77\%. From the success rates, it is also clear that the presence of distractor items or changes in lighting conditions, both of which are important real-world domain shifts, do not influence the performance of the keypoint detector. This shows that the randomizations in the synthetic data generation were effective in this regard.

The fold success rate is about 53\%. Examples of successful folds (and hence grasps) for various settings can be seen in Fig.~\ref{fig:head-figure}. The fold success rate is lower than the grasp success rate due to the limitations of scripted trajectories and single-arm manipulation: About 80\% of all fold failures are caused by corner misalignment as the corners bend during the execution of the fold motion ( see bottom row of  Fig.~\ref{fig:failed-fold-examples}). This could be resolved with bimanual manipulation or by enlarging the fingertips to better perform a line fold~\cite{Borras2020}. For towels that are already partially folded, the thickness and material properties can change, which makes the open-loop trajectory no longer suited to fold them (see top row example in Fig.~\ref{fig:failed-fold-examples}), explaining the remaining fold failures.

Most grasp failures are due to inaccurate or incomplete keypoint detection by the transferred neural network, examples of which can be seen in Fig.~\ref{fig:failed-keypoints}. These failures indicate there is still a reality gap that was not covered by the variations introduced in our synthetic dataset, requiring additional tuning to better match the target distribution. For the towels that contained non-uniform textures and were hence not explicitly modelled, the grasp success rate deteriorates to 48\% (see Table~\ref{table:evaluation-metrics}). 
The remaining grasp failures are due to some towels being very light, causing the fingers to push the cloth away instead of reaching underneath and grasping it due to a lack of friction.

Finally, as recommended in~\cite{benchmarking_cloth_manipulation},  we report the execution time of our robot system: A forward pass of the neural network is, even on CPU, negligibly fast ($\ll$ 1\;s) compared to the time it takes to execute the grasp and fold, which is about 20\;s.

\begin{table}

\centering
\begin{tabular}{rcccc}
\hline
    Setting & \multicolumn{2}{c}{In-distribution Towels} & \multicolumn{2}{c}{Out-of-distribution Towels}\\
 & Grasp & Fold & Grasp  & Fold  \\ \hline
Natural Light     & 17/22         & 14/22   & 9/18 & 8/18      \\
LED Light     & 16/22         & 10/22  & 8/18 & 8/18      \\
Distractors    & 18/22         & 11/22    & 9/18 & 4/18     \\ 
Total & 51/66 (77\%)  & 35/66 (53\%) & 26/54 (48\%) & 20/54 (37\%) \\
\hline
\end{tabular}
\newline
    \caption{Performance of the system for real-world robotic folding of towels using zero-shot transfer for the keypoint detector.}
    \label{table:evaluation-metrics}
\end{table}

\section{Conclusion and Future Work}
\label{section:conclusion}
In this paper, we used synthetic data to train a neural network to detect keypoints as a low-dimensional state representation for cloth folding. We transferred the detector without any finetuning (zero-shot) and measured its performance by using the keypoints for folding towels on a robot setup. The results indicate that using procedural data generation is a viable approach to training keypoint detectors for cloth manipulation. However, more extensive tuning is required to completely overcome the reality gap. In future work, we plan to further explore what factors matter to close this gap. This will enable increased performance and provide more principled guidelines for procedural data generation in general.

Additionally, the gap between the grasp and fold success rates indicates that not only the perception but also the manipulation of cloth remains a challenging task. In future work, we therefore aim to close the control loop to take the cloth properties into account and make the control more robust.


\section*{Acknowledgment}
The authors wish to thank the members of the \textit{Keypoints Gang}, in particular Rembert Daems and Peter De Roovere, for sharing many insights into Computer Vision and academic research in general. This research is supported by the Research Foundation Flanders (FWO) under Grant numbers 1S56022N (TL) and 1SD4421N (VDG).

\clearpage
\bibliographystyle{IEEEtran}
\bibliography{references.bib}
\end{document}